\documentclass[nohyperref]{article}

\usepackage{microtype}
\usepackage{graphicx}
\usepackage{subfigure}
\usepackage{booktabs} 

\usepackage{hyperref}

\usepackage[accepted]{icml2023}

\usepackage{amsmath}
\usepackage{amssymb}
\usepackage{mathtools}
\usepackage{amsthm}

\usepackage{listings}
\usepackage[capitalize,noabbrev]{cleveref}

\theoremstyle{plain}

\theoremstyle{definition}

\theoremstyle{remark}

\usepackage[textsize=tiny]{todonotes}

\icmltitlerunning{SNeRL: Semantic-aware Neural Radiance Fields for Reinforcement Learning}

\begin{document}

\twocolumn[
\icmltitle{SNeRL: Semantic-aware Neural Radiance Fields for Reinforcement Learning}



\icmlsetsymbol{equal}{*}

\begin{icmlauthorlist}
\icmlauthor{Dongseok Shim}{equal,yyy}
\icmlauthor{Seungjae Lee}{equal,comp,sch}
\icmlauthor{H. Jin Kim}{yyy,comp,sch}

\end{icmlauthorlist}

\icmlaffiliation{yyy}{Interdisciplinary Program in AI, Seoul National University}
\icmlaffiliation{comp}{Aerospace Engineering, Seoul National University}
\icmlaffiliation{sch}{ASRI, AIIS, Seoul National University}

\icmlcorrespondingauthor{H. Jin Kim}{hjinkim@snu.ac.kr}
\icmlkeywords{Machine Learning, ICML}

\vskip 0.3in
]



\printAffiliationsAndNotice{\icmlEqualContribution} 

\begin{abstract}
As previous representations for reinforcement learning cannot effectively incorporate a human-intuitive understanding of the 3D environment, they usually suffer from sub-optimal performances.
In this paper, we present Semantic-aware Neural Radiance Fields for Reinforcement Learning (SNeRL), which jointly optimizes semantic-aware neural radiance fields (NeRF) with a convolutional encoder to learn 3D-aware neural implicit representation from multi-view images.
We introduce 3D semantic and distilled feature fields in parallel to the RGB radiance fields in NeRF to learn semantic and object-centric representation for reinforcement learning.
SNeRL outperforms not only previous pixel-based representations but also recent 3D-aware representations both in model-free and model-based reinforcement learning.
\end{abstract}

\section{Introduction}

Developing agents that can achieve complex control tasks directly from image inputs has been a long-standing problem in reinforcement learning (RL). 
Previous works over the past few years have made notable progress in the data efficiency of learning visual control problems. 
The most challenging part of solving visual control tasks is obtaining the low-dimensional latent representations from high-dimensional observations. 
To this end, they pre-train the encoder in various ways such as unsupervised representation learning via image reconstruction using offline datasets \cite{finn2016deep, kulkarni2019unsupervised, islam2022discrete}, contrastive learning \cite{zhan2022learning}, reconstructing task information \cite{yang2021representation, yamada2022task}, and training multi-view consistency \cite{dwibedi2018learning}. 
 Other approaches utilize joint learning of auxiliary unsupervised tasks \cite{laskin2020curl, schwarzer2020data}, and data-augmented reinforcement learning \cite{laskin2020reinforcement, yarats2021mastering}.

\begin{figure}
    \centering
    \includegraphics[width=0.5\textwidth]{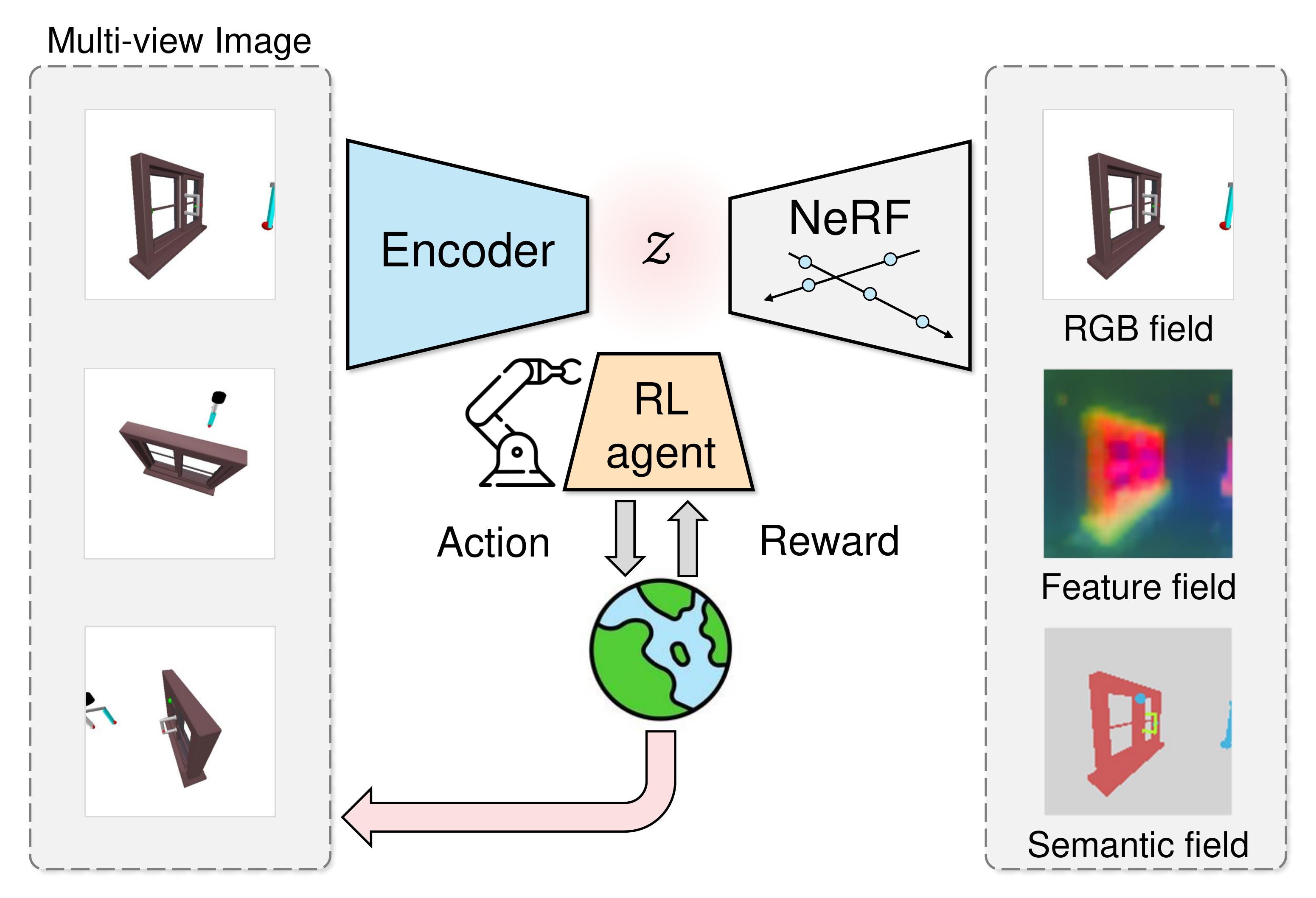}
    \vspace{-26pt}
    \caption{
    Semantic-aware NeRF for reinforcement learning.
    We present SNeRL, a reinforcement learning framework that learns 3D-aware representation with a convolutional encoder and semantic-aware NeRF decoder.
    The latent vectors from the encoder are propagated to the policy network to generate an action for RL agents.
    }
    \vspace{-22pt}
    \label{fig:thumbnail}
\end{figure}

While a number of works have been proposed to improve the data efficiency in visual control problems, the majority of the encoders trained from those methods have limited capability in obtaining 3D structural information and lack equivariance to 3D transformations.
Such limitations come from ignoring 3D structural information and learning visual representation from a single-view observation.

Recently, there have been attempts to consider 3D information of the environment in robot control and manipulation \cite{li20223d,driess2022reinforcement} by learning implicit spatial representation via neural radiance fields (NeRF) \cite{mildenhall2020nerf}.
They map pixel-level multi-view observations of a scene to a latent vector through an autoencoder structure, where the NeRF decoder provides 3D structured neural scene representation via RGB self-supervision for each view.

Even though the aforementioned pioneers achieved better performance compared to the previous RL algorithms with a single-view observation, they still did not take full advantage of 3D-aware representation learning.
It is because those methods only exploit RGB supervision to train NeRF, which makes it difficult for the encoder to learn $\textit{object-centric}$ or $\textit{semantic}$ representation for RL downstream tasks.
Although NeRF-RL \cite{driess2022reinforcement} proposes compositional NeRF to mitigate such limitations, the RL agents in NeRF-RL require object-individual masks during training and deployment to utilize semantic representations, which is quite unrealistic.

In this work, we propose Semantic-aware Neural Radiance Fields for Reinforcement Learning (\textbf{SNeRL}) which learns both 3D-aware semantic and geometric representation for RL agents in a 3-dimensional environment.
First of all, our proposed method learns 3D-aware semantic representation by predicting 3D semantic fields with ground truth labels. As a result, SNeRL enables downstream visual control tasks without object-individual masks and addresses the limitation of the prior work \cite{driess2022reinforcement}. 
Also, to capture further fine-grained features that could not be fully expressed in semantic fields and to take advantage of data-driven approaches, we employ an off-the-shelf feature descriptor \cite{caron2021emerging} as a teacher network and learn to predict feature fields via a distillation method such as \citet{kobayashi2022decomposing}.


We also introduce a multi-view adaptation of recent self-predictive representation learning \cite{chen2021exploring} as an auxiliary task which further improves the performance of SNeRL. In the proposed auxiliary task, SNeRL computes the target representation by utilizing the observations from different camera views in the same timestep to learn spatially consistent representation.


Our proposed SNeRL outperforms not only the previous single-view representation learning algorithms for RL \cite{laskin2020curl, yarats2021mastering} but also the state-of-the-art method with multi-view observations \cite{driess2022reinforcement} in four different visual control tasks. 

To sum up, our contribution can be summarized as follows:

\begin{itemize}
    \item We present SNeRL, a framework that utilizes NeRF with semantic and distilled feature fields to learn 3D-aware semantic representation for reinforcement learning. 
    \item We validate the effectiveness of SNeRL both with model-free and model-based methods. To the best of our knowledge, SNeRL is the first work that leverages semantic-aware representations without object masks in RL downstream tasks. Also, this is the first study to utilize 3D-aware representations to model-based RL.
    \item The proposed SNeRL outperforms the previous single and multi-view image-based RL algorithms in four different 3D environments from Meta-world. In addition, auxiliary self-predictive representation learning with multi-view observations proposed for spatially consistent representation can enable further improvements.
\end{itemize}

\section{Related Work}
\subsection{3D Scene Representation Learning}
To learn 3D-aware representation from a single view image, the previous methods exploit standard convolutional autoencoder architecture conditioned by the camera poses, which generates scenes from arbitrary views with either deterministic \cite{tatarchenko2016multi, worrall2017interpretable} or stochastic \cite{eslami2018neural} latent vectors.
Recently, neural radiance fields (NeRF) have achieved an exceptional progress in understanding 3D scenes and synthesizing novel views.
Following, some approaches propose latent-conditioned NeRF \cite{martin2021nerf, yu2021pixelnerf, wang2021ibrnet}, but the major objective of the aforementioned methods is improving the quality of synthesized images rather than extracting time-variant latent vectors with 3D dynamic scene understanding from multi-view inputs.
In this paper, we leverage the autoencoder with convolutional encoder and NeRF-style decoder \cite{li20223d, driess2022reinforcement} so that the encoder can extract 3D-aware representation from multi-view inputs for RL downstream tasks.

\subsection{Representation Learning for RL}
The RL frameworks with image inputs typically have an encoder, which maps high-dimensional observations to a low-dimensional latent vector. 
RL agent is trained over the latent state space to maximize its objective functions, e.g., the total discounted reward for each episode. While a number of works have made significant advancements, it still remains a challenging open problem.

To address the sample inefficiency of image-based RL, prior works adopt various data-augmentation techniques \cite{laskin2020reinforcement, yarats2021mastering}, contrastive learning with data augmentation \cite{laskin2020curl, schwarzer2020data, stooke2021decoupling, liu2021behavior, zhan2022learning}, representation learning from image reconstruction \cite{islam2022discrete, kulkarni2019unsupervised}, or task information reconstruction \cite{yang2021representation, yamada2022task}. 
Other approaches propose to capture the relations between multi-view data \cite{dwibedi2018learning, kinose2022multi, sermanet2018time} or keypoints \cite{manuelli2020keypoints}. 
There are also some approaches leveraging transition sequence data \cite{hansen2020self, you2022integrating}, or pre-training with offline image-based RL \cite{wang2022vrl3}. Unfortunately, these works have limited capability in learning 3D-structural information and could not obtain an intuitive understanding of the 3D environments that humans have because of the 2D bias that 2D convolutional neural networks have.

In recent, there have been attempts to learn the 3D structure of the real world \cite{li20223d, driess2022reinforcement}. 
\citet{li20223d} firstly proposes autoencoder with convolutional encoder and NeRF \cite{mildenhall2020nerf} decoder to control the visuomotor with learned dynamics model and model-predictive control (MPC).
Following, NeRF-RL \cite{driess2022reinforcement} extends the prior study and firstly introduces NeRF-based architecture to the general model-free RL framework.
However, they could not learn semantic features due to the limited RGB supervision with na\"ive NeRF.
To learn object-centric representation only with RGB supervision, NeRF-RL presents compositional NeRF with object-individual masks, but requiring masks during the deployment of RL agents seems to be a strong assumption.

In this paper, we propose SNeRL which learns both geometric and semantic information with RGB, semantic, and distilled feature supervision for RL downstream tasks without any object masks during the inference phase.

%

\section{Preliminaries}
\subsection{Neural Radiance Fields}
The concept of neural radiance fields (NeRF) \cite{mildenhall2020nerf} is to represent the 3D scene with learnable and continuous volumetric fields $f_\theta$.
Specifically, at any 3D world coordinate $\mathbf{x} \in \mathbb{R}^{3}$ and unit viewing direction $\mathbf{d} \in \mathbb{R}^{3}$, $f_\theta$ estimates the differntiable volume density $\sigma$ and RGB color $\mathbf{c}$: $f_\theta(\mathbf{x}, \mathbf{d})  = (\sigma, \mathbf{c})$.
Let the camera ray of the pixel in the camera coordinate be $\mathbf{r}=\mathbf{o}+t\mathbf{d}$, where $\mathbf{o}$ indicates the camera origin.
The corresponding pixel value from an arbitrary view can be rendered through volumetric radiance fields as: 
\begin{equation}
    C(\mathbf{r}) = \int^{t_{f}}_{t_{n}}T(t)\sigma(t)\mathbf{c}(t)dt
\end{equation}
where $ T(t)=\mathrm{exp}(-\int^{t}_{t_{n}}\sigma(s)ds)$ and $t_n$ and $t_f$ indicate pre-defined lower and upper bound of the depth respectively.

Then, $f_\theta$, which is usually formulated with MLP, is optimized through pixel-wise RGB supervision from multiple views as:
\begin{equation}
    \mathcal{L} = \sum_{i,j}||\hat{C}(\mathbf{r}_{i,j}) - C(\mathbf{r}_{i,j})||_2^2,
\end{equation}
where $\mathbf{r}_{i,j}$ indicates ray $j$ from images of $i^{th}$ view.
$\hat{C}$ and $C$ represents the rendered volumetric fields into 2D image and ground truth pixel value respectively.
\subsection{Reinforcement Learning}

We consider a finite-horizon Markov Decision Process (MDP) $\mathcal{M}=(\mathcal{O}, \mathcal{A}, \mathcal{T}, \mathcal{R}, \gamma)$, where $\mathcal{O}$ denotes the high-dimensional observation space (image pixels), $\mathcal{A}$ the action space, $\mathcal{T}(o'|o,a)$ the transition dynamics $(o,o'\in\mathcal{O}, a\in\mathcal{A})$, $\mathcal{R}:\mathcal{O}\times\mathcal{A} \rightarrow \mathbb{R}$ the reward function, and $\gamma\in[0,1)$ the discount factor. Following the general idea of learning RL downstream tasks with pre-trained scene representations, we 
consider an encoder $\Omega:\mathcal{O}\rightarrow\mathcal{Z}$ that maps and high-dimensional observation $o\in\mathcal{O}$ to a low-dimensional latent state $z\in\mathcal{Z}$ on which an RL agent operates. To learn how to succeed in downstream tasks, the RL policy $\pi_\theta (a\in\mathcal{A}|z=\Omega(o)) $ maximizes the total discounted reward $\sum_{t=0}^{H-1}=\gamma^t \mathcal{R}(o_t,a_t)$ of trajectories $\tau_i=(z_0, o_0, ..., z_H, o_H)_i$.

\section{Method}
\begin{figure*}[t!]
    \centering
    \includegraphics[width=1.\textwidth]{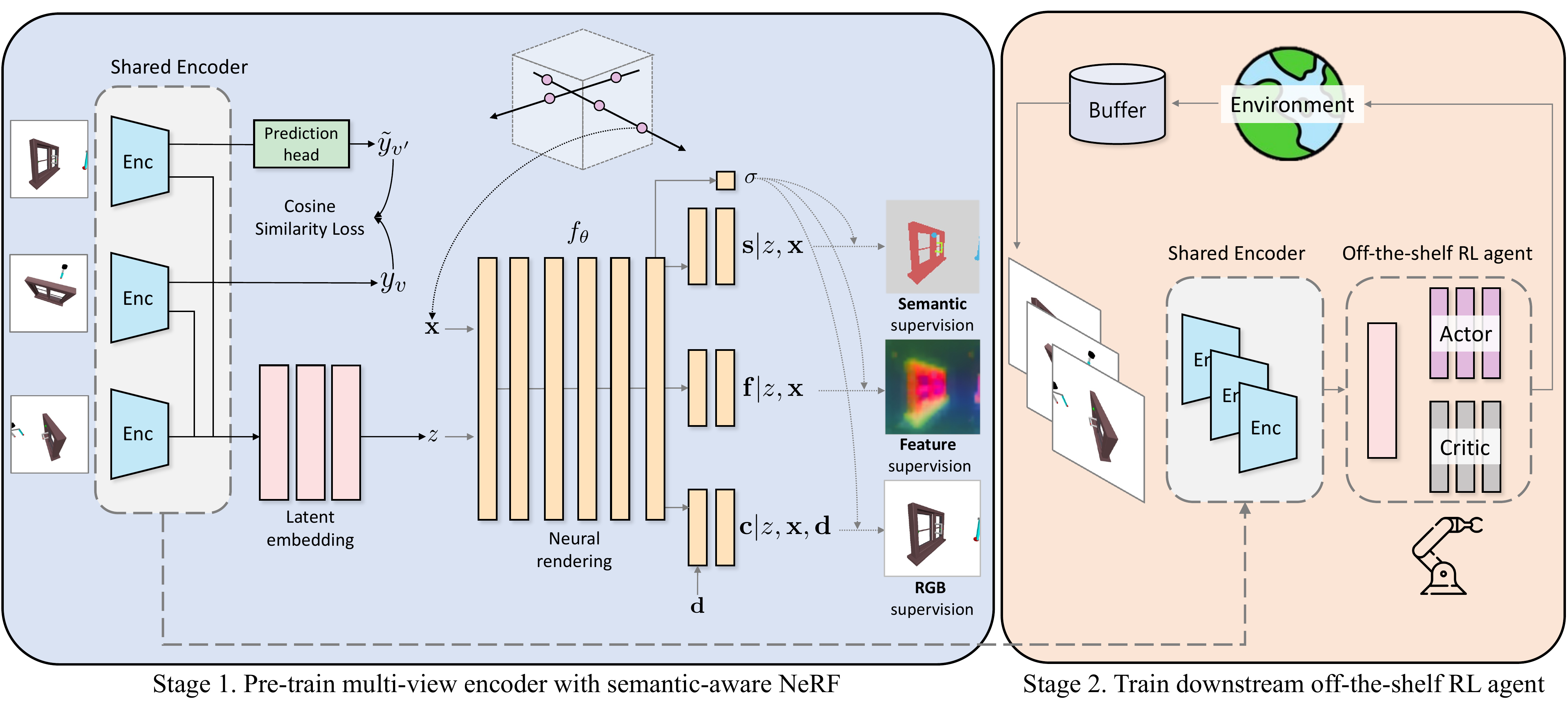}
    \vspace{-12pt}
    \caption{
    SNeRL Overview. 
    SNeRL consists of two stages, which are pre-training NeRF-based autoencoder and fine-tuning to the downstream RL tasks, respectively.
    With observations from three different camera views, an encoder produces a single latent vector $z$, and
    a decoder with neural rendering function $f_\theta$ takes the position $\mathbf{x}$, viewing direction $\mathbf{d}$ in the 3D coordinates and $z$ as inputs to synthesize three different fields in the arbitrary views.
    An auxiliary multi-view self-prediction loss is applied to enable view-invariant representation.
    Then, the encoder and the decoder are jointly optimized in a supervised manner with an offline dataset.
    The pre-trained encoder is utilized as a feature extractor to train the policy with off-the-shelf RL algorithms. 
    }
    \vspace{-12pt}
    \label{fig:overview}
    
\end{figure*}

In this section, we demonstrate the details of SNeRL which consists of a multi-view convolutional image encoder and a latent-conditioned NeRF decoder to learn the 3D-aware representation.
Compared to the previous method \cite{driess2022reinforcement} which also proposes NeRF supervision for RL, SNeRL is capable of extracting $\textit{object-centric}$ or $\textit{semantic}$ representation without any object-individual masks during deployment.
The pre-trained image encoder is exploited as a feature extractor for downstream RL tasks, and the overview of SNeRL framework is depicted in Figure \ref{fig:overview}.

\subsection{Multi-view Encoder}
Similar to \citet{li20223d}, we adopt the multi-view encoder $\Omega$ which fuses the observations from multiple camera views together to learn a single latent vector $z$ for RL tasks.
The encoder takes the pixel-level observations $o^{i} \in \mathbb{R}^{H\times W\times3}$, and the corresponding camera projection matrices $K^{i} \in \mathbb{R}^{3\times4}$ captured from $V$ different camera view as inputs, i.e., $i=1\cdots V$.
To generate $z\in\mathcal{Z}$ from the inputs, a convolutional network $E_{\mathrm{CNN}}$ first extracts viewpoint-invariant features from each image.
The features from different camera views are channel-wise concatenated with their corresponding (flattened) camera projection matrices to reflect the viewpoint information to the following feature vectors.
Then, the concatenated vectors are passed through MLP layers, $g_{\mathrm{MLP}}$, to produce mid-level viewpoint-aware encodings.
Lastly, the feature encodings from different camera views are averaged to generate a single encoding, and the averaged feature encoding is projected to the latent space $\mathcal{Z}$ with the latent encoder $h_{\mathrm{MLP}}$ as follows:
\begin{equation}
\begin{split}
    z =& \,\Omega (o^{1:V}, K^{1:V})\\
      =& \,h_{\mathrm{MLP}}(\frac{1}{V}\sum_{i=1}^{V}g_{\mathrm{MLP}}(E_{\mathrm{CNN}}(o^{i}), K^{i}))
\end{split}
\end{equation}
\subsection{Semantic-aware NeRF Decoder}
To inject 3D structural information into the latent vector $z$, we leverage a latent-conditioned NeRF architecture \cite{yu2021pixelnerf, martin2021nerf, wang2021ibrnet} for the decoder.
The difference between previous latent-conditioned NeRF and our proposed SNeRL is that the neural rendering function $f_\theta$ from SNeRL not only synthesizes novel views with RGB pixel value $\mathbf{c}$ but also with the semantic label $\mathbf{s}$ \cite{Zhi:etal:ICCV2021, fu2022panoptic, kundu2022panoptic} and high-dimensional distilled features $\mathbf{f}$ from the large-scale teacher network \cite{kobayashi2022decomposing} as follows:
\begin{equation}
    \mathbf{c} = f_\theta (z, \mathbf{x}, \mathbf{d}),\,\, \mathbf{s},\, \mathbf{f} = f_\theta (z, \mathbf{x})     
\end{equation}
By estimating three different radiance fields (semantic, feature, and RGB), the latent vector $z$ is jointly optimized to learn the geometric and semantic representations of the 3D environment.
Unlike RGB value $\mathbf{c}$ which is dependent on both the position $\mathbf{x}$ and the viewing direction $\mathbf{d}$, we formulate the semantic label and distilled feature to be invariant to the viewing direction $\mathbf{d}$ because the inherent properties of the scene or the object do not change according to the direction of the camera ray.

As SNeRL predicts three different fields, RGB, semantic, and distilled feature, by adding field-wise branches, they share the neural rendering function $f_\theta$ until estimating the density $\sigma$.
It indicates that three radiance fields have the same accumulated transmittance $T(t)$ at depth $t \in [t_{n}, t_f]$ along the ray $\mathbf{r} = \mathbf{o} + t\mathbf{d}$ as
\begin{equation}
    T(t) = \mathrm{exp}(-\int^{t}_{t_n}\sigma(\mathbf{r}(s))ds).
\end{equation}

For rendering the RGB field, we follow the same training framework as general latent-conditioned NeRF \cite{yu2021pixelnerf, martin2021nerf, wang2021ibrnet}, which optimizes the neural rendering function $f_\theta$ via pixel-wise RGB supervision.
RGB supervision enables the encoder to extract geometric features from the observed environment by learning the RGB and density distribution in the 3-dimensional space.
The rendered pixel value $\hat{C}(\mathbf{r})$ can be calculated as
\begin{equation}    
    \hat{C}(\mathbf{r}) = \int^{t_{f}}_{t_{n}}T(t)\sigma(\mathbf{r})\mathbf{c}(\mathbf{r}, \mathbf{d})dt,
\end{equation}
and the loss function for RGB field, $\mathcal{L}_{\mathrm{RGB}}$, can be formulated with simple L2 loss between the rendered $\hat{C}(\mathbf{r})$ and the ground truth pixel colors $C(\mathbf{r})$,
\begin{equation}
    \mathcal{L}_{\mathrm{RGB}} = \sum_{i,j}||\hat{C}(\mathbf{r}_{i,j}) - C(\mathbf{r}_{i,j})||_2^2,
\end{equation}
where $\mathbf{r}_{i,j}$ indicates the camera ray $j$ from the observation $i$, $o^{i}$.

Unfortunately, optimizing the encoder only with only an RGB reconstruction is difficult to capture the $\textit{semantic}$ or $\textit{object-centric}$ properties of the 3D scene, which are crucial for downstream RL tasks.
Therefore, we extend NeRF-based decoder by appending additional branches before injecting the viewing direction $\mathbf{d}$ into the rendering function, $f_\theta$ for semantic segmentation.
The rendered semantic labels $\hat{S}(\mathbf{r})$ can be calculated as
\begin{equation}
    \hat{S}(\mathbf{r}) = \int^{t_{f}}_{t_{n}}T(t)\sigma(\mathbf{r})\mathbf{s}(\mathbf{r})dt
\end{equation}
and the loss function for semantic field, $\mathcal{L}_{seg}$, can be formulated with the standard cross entropy loss,
\begin{equation}
    \mathcal{L}_{sem} = -\sum_{i,j}\sum_{l=1}^L S^{l}(\mathbf{r}_{i,j})\mathrm{log}\,\hat{S}^{l}(\mathbf{r}_{i,j}),
\end{equation}
where $\hat{S}^{l}$ and $S^{l}$ denote the probability of the ray $j$ in observation $i$ belonging to the class $l$ and its corresponding ground-truth semantic labels, respectively.

To capture further fine-grained features that could not be fully expressed in semantic fields, SNeRL also synthesizes distilled feature fields (\citet{kobayashi2022decomposing}) that predict the output of a pre-trained feature descriptor in a knowledge-distillation manner \cite{hinton2015distilling}.
It is well known from prior literature \cite{caron2021emerging} that Vision Transformer (ViT) \cite{dosovitskiy2020image} trained in a self-supervised manner, e.g. DINO \cite{caron2021emerging}, can work as an excellent feature descriptor which explicitly represents the scene layouts such as object boundaries.
Since the output of the ViT feature descriptor contains high-dimensional information with different values in all pixels depending on the geometric relationship or semantic meaning, the pre-trained ViT becomes a good feature descriptor with another advantage from the semantic label.

Therefore, we take advantage of such benefits to the NeRF-based decoder so that the latent vector $z$ learns high-level information by distilling the knowledge from ViT teacher network which cannot be learned via ground-truth semantic supervision.
The distilled feature fields can be rendered as
\begin{equation}
    \hat{F}(\mathbf{r}) = \int^{t_{f}}_{t_{n}}T(t)\sigma(\mathbf{r})\mathbf{f}(\mathbf{r})dt.
\end{equation}
The loss function for distilled feature field, $\mathcal{L}_{feat}$, is formulated by penalizing the difference between the rendered features $\hat{F}(\mathbf{r})$ and the outputs of ViT feature descriptor $F(o, \mathbf{r})$ as
\begin{equation}
    \mathcal{L}_{feat} = \sum_{i,j}||\hat{F}(\mathbf{r}_{i,j}) - F(o^{i}, \mathbf{r}_{i,j})||_1.
\end{equation}

Finally, the total loss function $\mathcal{L}$ for jointly optimizing the multi-view encoder and NeRF-based decoder can be formulated as the linear combination of aforementioned losses as:
\begin{equation}
    \mathcal{L} = \mathcal{L}_{\mathrm{RGB}} + \lambda_{sem}\mathcal{L}_{sem} + \lambda_{feat}\mathcal{L}_{feat}
\end{equation}
where $\lambda_{sem}$ and $\lambda_{feat}$ are set to 0.004 and 0.04, respectively, to balance the losses \cite{Zhi:etal:ICCV2021, kobayashi2022decomposing}.
After training, the multi-view encoder $\Omega$ is exploited as a 3D structural and semantic feature extractor for any off-the-shelf downstream RL algorithms.

\subsection{Multi-view Self Predictive Representation} \label{sec:4.3}
We additionally enforce the multi-view self-predictive loss to the latent vector $z$ to ensure that the encoder learns the viewpoint-invariant representation with observations from the same scene.
The randomly sampled observations from two different camera pose, $o^{1}$ and $o^{2}$, are processed by the convolutional feature extractor, $E_{\mathrm{CNN}}$, and the weights of the feature extractor are shared between two inputs.
A feature from one view, $z_1$, is mapped with a prediction network, $h_{pred}$, to match it to the feature from the other view, $z_2$.
We formulate the self-predictive loss function $D$ with negative cosine similarity as follows:
\begin{equation}
    \mathcal{D}(p_1, z_2) = -\frac{p_1}{{||p_{1}||_{2}}}\cdot\frac{z_2}{{||z_{2}||_{2}}},
\end{equation}
where $p_1$ and $z_2$ indicate two output vectors, $p_1\triangleq h_{pred}(E_{\mathrm{CNN}}(o^{1}))$ and $z_2\triangleq E_{\mathrm{CNN}}(o^{2})$, respectively.
We assume that $z_2$ is constant and the encoder $E_{\mathrm{CNN}}$ only receives the gradient from $p_1$ following \citet{chen2021exploring}.

The symmetrized auxiliary representation loss function can be formulated as follows:
\begin{equation}
    \mathcal{L}_{\mathrm{aux}} = \frac{1}{2}\mathcal{D}(p_1, z_2) + \frac{1}{2}\mathcal{D}(p_2, z_1).
\end{equation}

\section{Experiments}
\begin{figure*}
    \centering
    \includegraphics[width=0.9\textwidth]{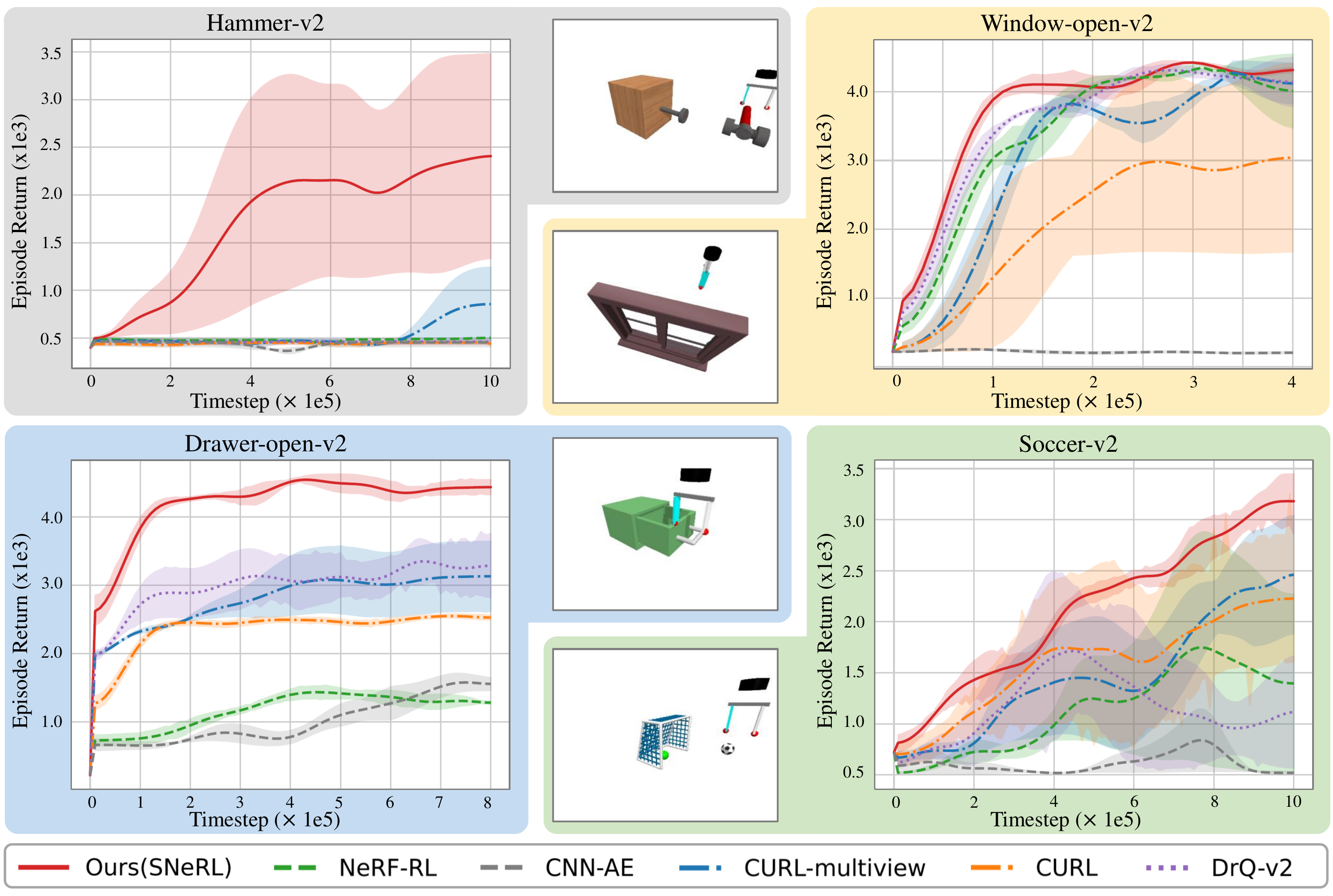}
    \vspace{-0.1in}
    \caption{
    Episode returns of the evaluation results. Shading indicates a standard deviation across 4 seeds. The curves are not visible in the  Hammer-v2 environment as they overlap each other. Note that SNeRL in this figure is obtained without auxiliary loss in section \ref{sec:4.3} (multi-view self-predictive presentation), which could enable further improvements in some environments.
    }
    \vspace{-0.15in}
    \label{fig:main_result}
    
\end{figure*}
In this section, we demonstrate several experiments on the 3-dimensional environments to explore the effectiveness of SNeRL compared to existing state-of-the-art RL algorithms both in model-free and model-based settings.
We fix the downstream RL algorithms and adopt Soft Actor-Critic \cite{haarnoja2018soft} in the model-free setting and Dreamer \cite{hafner2019dream} in the model-based setting for SNeRL and all the baselines for a fair evaluation.

\paragraph{Environments.} 
We evaluate SNeRL on four visual control environments based on the MuJoCo \cite{todorov2012mujoco}, including some complex control tasks that require clever use of interactions between the objects to obtain high rewards. 
All the tasks are performed by a simulated Sawyer robot which has a single arm and gripper in hand (4-DoF). The action space of the Sawyer robot consists of the position (x,y,z) of the end-effector and gripper control (open/close). 
The agent takes 128x128 images from three different camera views as pixel-level inputs and receives dense rewards from the environment provided by Meta-world \cite{yu2020meta}.

\begin{itemize}
    \item Window-open-v2 : This environment involves the Sawyer robot opening a sliding window with a handle. The initial state of the robotic hand is [0, 0.4, 0.2] and the robot receives rewards for pushing the handle and opening a window located in [-0.1, 0.785, 0.16].
    \item Hammer-v2 : The Sawyer robot is supposed to grasp the handle of the hammer, which is generated in a random position, and hit the head of the nail to drive it. The initial state of the robotic hand is generated randomly in $\{(x,y,z)|-0.5\leq x\leq 0.5, 0.4\leq y\leq 1, 0.05\leq z\leq 0.5\}$. The robot receives rewards for picking up the hammer and inserting the nail into a piece of wood.
    
    \item Drawer-open-v2 : The Sawyer robot is supposed to open a drawer by holding the handle of the drawer and pulling it. The initial state of the robotic hand is the same as Hammer-v2. The robot receives rewards for opening a drawer.
    
    \item Soccer-v2 : In this task, the Sawyer robot tries to score by pushing a soccer ball that is generated in a random position. The initial state of the robotic hand is the same as Hammer-v2. The robot receives rewards for touching the soccer ball and putting it into the net.
\end{itemize}

We refer to Meta-world \cite{yu2020meta} for more details including the reward function and the range of the random positions.

\begin{figure*}
    \centering
    \includegraphics[width=0.9\textwidth]{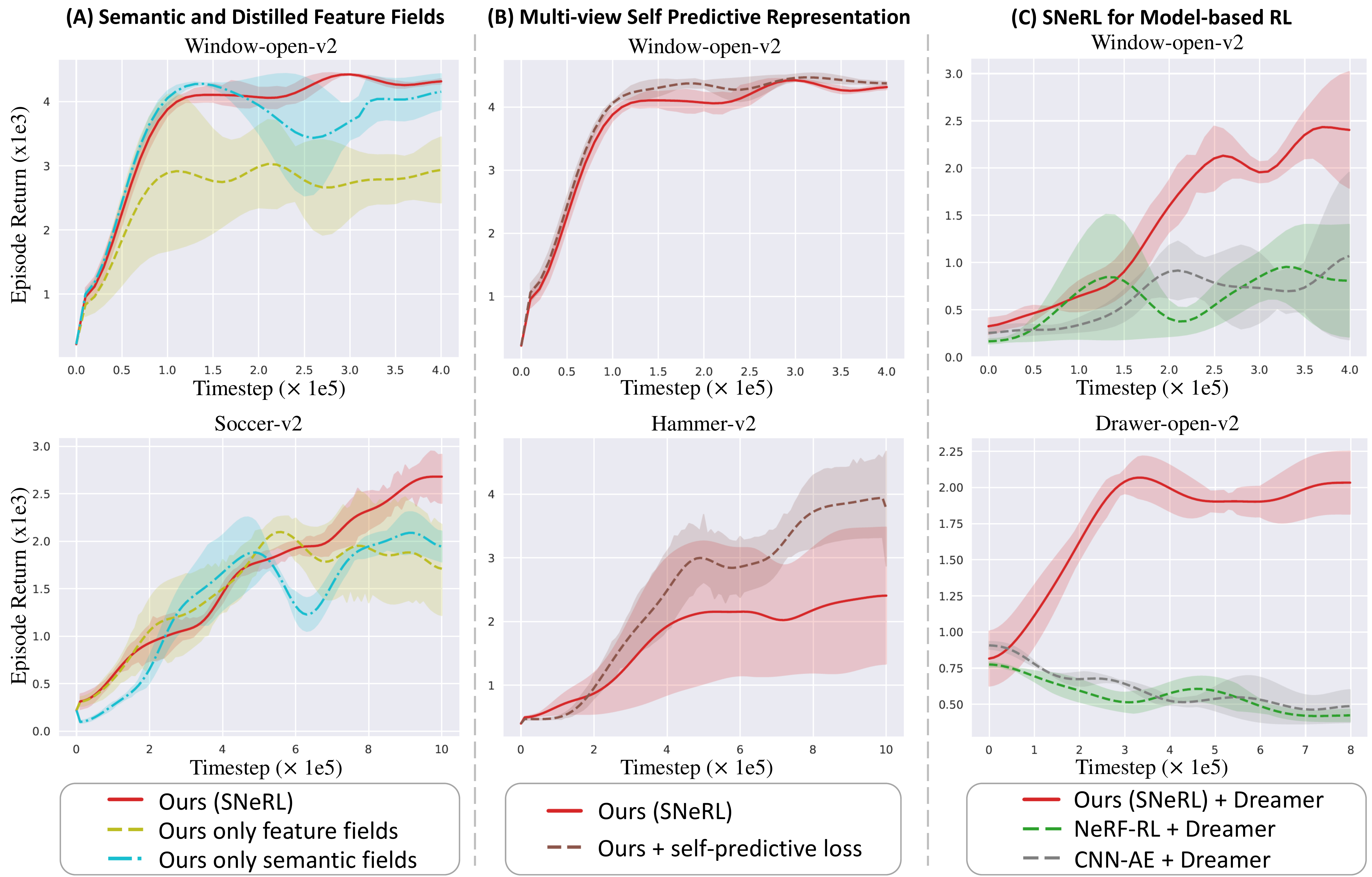}
    \vspace{-0.07in}
    \caption{
    Ablation study. (a): SNeRL with both semantic and feature supervision shows higher performance than the case where only one of the two is applied. Also, in relatively simple environments, using only one of the two could be enough to improve the performance of the prior work. (b): Additional multi-view self-predictive loss can further improve SNeRL in some environments. (c): Learned representations via SNeRL can also be adopted in model-based RL.
    }
    \vspace{-0.17in}
    \label{fig:ablation}
\end{figure*}

\paragraph{Baselines.}
We compare SNeRL to several state-of-the-art visual RL methods and a 3D-aware RL method, which are briefly described below. 
\textbf{DrQ-v2} \cite{yarats2021mastering} is an improved version of DrQ \cite{yarats2020image}, which solves visual control tasks with data augmentation and scheduled exploration noise. 
\textbf{CURL} \cite{laskin2020curl} trains RL agents with an auxiliary contrastive loss which ensures that the embeddings for data-augmented versions of observations match. \textbf{CURL-multiview} is a multi-view adaptation of CURL, which utilize 3 different camera views and has a CNN encoder with the same structure as that of SNeRL. 
\textbf{CNN-AE} uses a standard CNN autoencoder (instead of NeRF decoder) to pre-train an encoder using the reconstruction loss proposed in \citet{finn2016deep}. 
\textbf{NeRF-RL} \cite{driess2022reinforcement} pre-trains an autoencoder with convolutional encoder and na\"ive NeRF-style decoder, without semantic and feature supervision. 

We note that learning downstream RL tasks in CNN-AE and NeRF-RL shares the identical method as SNeRL, and they use the same offline dataset collected by random actions and the policies provided by Meta-world (half-and-half mixed). We refer the reader to Appendix \ref{sec:B.2} for the experiments on other datasets.
Also, all the multi-view methods (CURL-multiview, CNN-AE, NeRF-RL, SNeRL) receive the same observations and do not receive per-object masks from the environment. 
For the rest of the baselines which operate on a single view, we choose a single camera position from which the states of each object can be observed clearly.

\subsection{Experiment Result}\label{sec:5.1}
Figure \ref{fig:main_result} shows the episode returns of SNeRL and baselines in 4 different visual control tasks. Thanks to the learned object-centric representation via semantic and distilled feature supervision, SNeRL consistently outperforms state-of-the-art visual RL methods and the prior 3D-aware RL method (NeRF-RL) in terms of data efficiency and performance.


Specifically, the contrastive baselines (CURL, CURL-multiview) and DrQ-v2 could not achieve high returns in the difficult environments (soccer and hammer) even though some of them succeed in the relatively easy environment (window). 
The results also show that pre-training CNN via na\"ive reconstruction loss (CNN-AE) with offline data could not succeed in the environments at all. 
These imply that extracting not only the 3D-aware geometric but also the $\textit{object-centric}$ and $\textit{semantic}$ information from multi-view observations is critical for RL performances.

Interestingly, we observe that pre-training a NeRF-based autoencoder only with RGB supervision (NeRF-RL) is not sufficient to learn the features for RL downstream tasks, and it can not outperform multi-view adaptation of the visual RL method with contrastive loss (CURL-multiview). 
These are contrary to the results reported in the prior work \cite{driess2022reinforcement}, which we analyze as follows: the environments we adopted in this work are more challenging compared to those of \citet{driess2022reinforcement}, which consist of simple-shaped objects with primary colors. 
Therefore, it is relatively difficult to obtain semantic information simply using RGB supervision. 
Thus, leveraging a semantic-aware NeRF decoder is required to extract the features for better performances in RL downstream tasks in the practical use of 3D-aware RL, which is consistent with our analysis.

\begin{figure*}[t]
    \centering
    \includegraphics[width=0.9\textwidth]{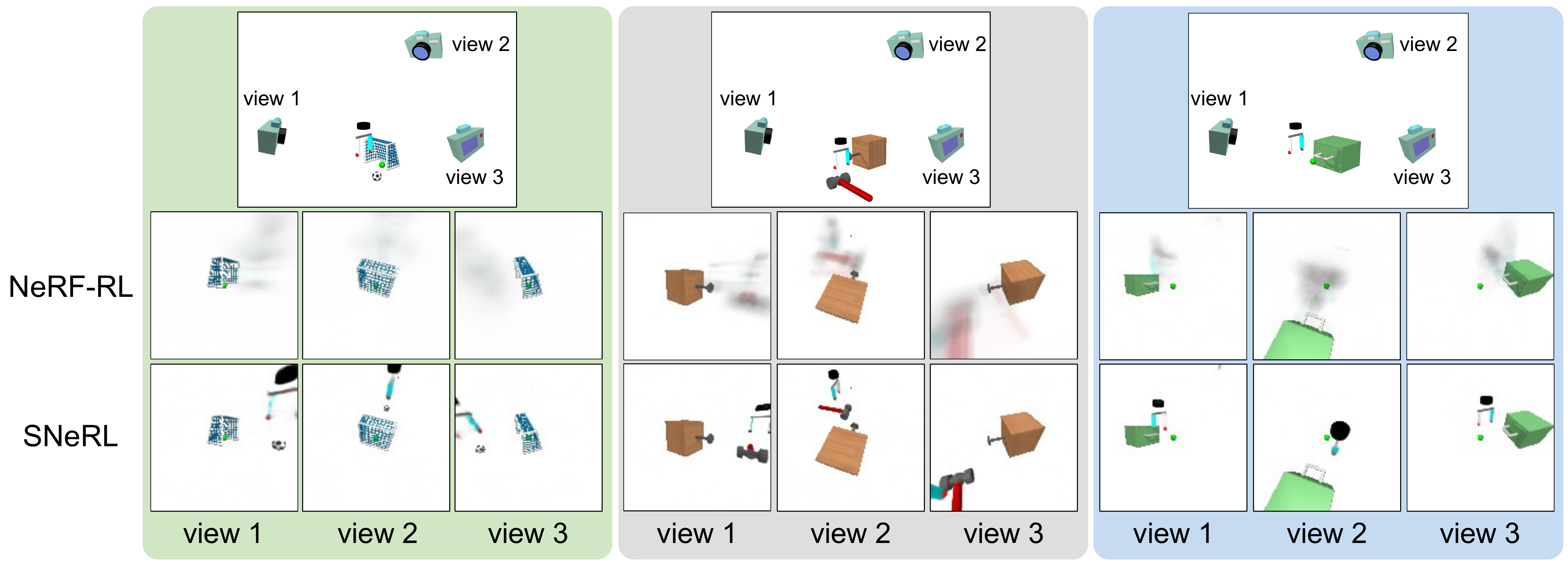}
    \vspace{-8pt}
    \caption{Qualitative results on the image reconstruction in 3 different camera views via neural rendering.
    The synthesized images from SNeRL achieve better fidelity compared to NeRF-RL in several environments.
    }
    \vspace{-12pt}
    \label{fig:recon}
    
\end{figure*}

\subsection{Ablation Study}
\paragraph{Semantic and Distilled Feature Fields.}
To validate how each semantic-aware radiance field leveraged in SNeRL (semantic and distilled feature fields) contributes to the downstream RL performances, we evaluate its two ablated variants without semantic and feature supervision, respectively.
As shown in Figure \ref{fig:ablation}(a), SNeRL, which takes advantage of both semantic and feature supervision from ground-truth labels and a ViT-based feature descriptor, achieves the best performance compared to all the ablated models.
We observe that the performance gap between SNeRL and the ablated models depends on the environment, as semantic labels are sufficient to learn semantic information in a relatively simple environment (window), but it requires both supervision in a complex environment (soccer).

\paragraph{Multi-view Representation Learning.}
We also introduce auxiliary representation learning which is suitable for multi-view observations with self-predictive loss.
To demonstrate its effectiveness on downstream RL tasks, we evaluate two different models, which are SNeRL with and without multi-view self-predictive representation learning.
By enforcing the latent vector to be invariant to the viewpoint of the observation, we report that the proposed representation learning improves the RL agent's performance in some environments as shown in Figure \ref{fig:ablation}(b).


\subsection{Image Reconstruction via Neural Rendering}
Even though we only leverage the convolutional encoder for downstream RL tasks, we compare the image rendering performance of NeRF-based decoders from SNeRL and NeRF-RL to explore the relationship between the synthesized image quality and RL performances.
As NeRF originally aims to synthesize images of arbitrary camera views from a static scene, NeRF-RL, which also trains the volumetric field with a sole RGB supervision, cannot reconstruct dynamic objects, e.g. the robot arm, in input images without semantic information.
On the other hand, SNeRL, which utilizes semantic labels and feature outputs from the ViT teacher network as additional supervision signals, not only achieves better RL performance but also well represents the dynamic scene and produces high-fidelity rendering outputs as shown in Figure \ref{fig:recon}.

\subsection{SNeRL for Model-based RL}
In this section, we evaluate whether the learned representation via SNeRL can also be adopted in off-the-shelf model-based reinforcement learning algorithms, which train a world model to characterize the environment and conduct planning over the learned model. We adopt Dreamer \cite{hafner2019dream} as a downstream model-based RL agent and replace the encoder of the representation model with our pre-trained encoder. Refer to Appendix \ref{sec:A} for additional implementation details and an architectural overview.

Our results are shown in Figure \ref{fig:ablation}(c). We observe that learning model-based RL with the pre-trained encoder of SNeRL outperforms the pre-trained weights of the prior 3D-aware RL method (NeRF-RL) and na\"ive CNN autoencoder. 
This empirical evidence is consistent with the case of model-free RL in section \ref{sec:5.1}, indicating that the proposed method allows the encoder to learn representations that are important for general off-the-shelf RL agents.


\section{Conclusion}
In this paper, we present SNeRL, a semantic-aware radiance field for RL, that outperforms existing representation learning methods for RL algorithms across four different 3-dimensional environments.
SNeRL leverages semantic and distilled feature supervision with latent condition NeRF autoencoders as well as RGB supervision to enable image encoders to express 3D-aware geometric and semantic representation on downstream RL tasks.
We also propose a multi-view self-predictive loss as an auxiliary representation learning to force latent vectors to be viewpoint invariant.
Finally, we verify that SNeRL is effective in both model-free and model-based RL algorithms.

\paragraph{Limitations.} 
Despite these improvements, SNeRL inherits the limitations of the prior 3D-aware RL method. 
First of all, SNeRL requires multi-view offline data, and collecting an offline dataset covering the state space in some complex control tasks might be challenging. 
Also, our method uses a NeRF decoder that consumes more computational budget than CNN, so there might be limitations in extending our method to an online setup which trains the encoder concurrently with RL agents.

\section{Acknowledgement}
This research was supported by Institute of Information \& communications Technology Planning \& Evaluation (IITP) grant funded by the Korea government(MSIT) [NO.2021-0-01343, Artificial Intelligence Graduate School Program (Seoul National University)]. Also, this research was supported by Unmanned Vehicles Core Technology Research and Development Program through the National Research  Foundation of Korea(NRF) and Unmanned Vehicle Advanced Research Center(UVARC) funded by the Ministry of Science and ICT, the Republic of Korea(NRF-2020M3C1C1A01086411). Seungjae Lee would like to acknowledge financial support from Hyundai Motor Chung Mong-Koo Foundation.
\bibliography{example_paper}
\bibliographystyle{icml2023}

\newpage
\appendix
\onecolumn
\section{Algorithms}\label{sec:A}

\subsection{Model-free RL : Soft Actor-Critic}

In this project, we adopt Soft Actor-Critic algorithm \cite{haarnoja2018soft} (SAC) in all the experiments of the model-free downstream RL. SAC optimizes stochastic policies to maximize both the expected trajectory returns and the expected entropy of the actions. Although SAC shows promising performances on a range of control tasks over continuous action spaces including many benchmark tasks, it fails or suffers from data inefficiency in some visual control tasks. To train actor network $\pi_\phi$ and critic networks $Q_{\theta_1}, Q_{\theta_2}$ SAC algorithm minimizes the following objective functions
\begin{equation}
    J_\pi(\phi) = \mathbb{E}_{{s_t}\sim \mathcal{D}} \big[ \mathbb{E}_{a_t\sim\pi_\phi}[\alpha\log(\pi_\phi(a_t|s_t))-\min_{i=1,2}Q_{\theta_i}(s_t,a_t)]\big]
\end{equation}

\begin{equation}\label{eq:16}
    J_Q(\theta_i) = \mathbb{E}_{{s_t,a_t,s_{t+1},r}\sim \mathcal{D}, a_{t+1}\sim\pi_\phi(s_{t+1})} \bigg[\frac{1}{2}(Q_{\theta_i}(s_t,a_t)-(r+\gamma\mathcal{T}))^2\bigg],
\end{equation}

where $\mathcal{D}$ denotes the replay buffer, $\alpha$ the temperature hyperparameter, and $\gamma$ the discount factor. The target value $\mathcal{T}$ in Eq. \ref{eq:16} is 
\begin{equation}
    \mathcal{T} = \min_{i=1,2}Q_{\hat{\theta}_i}(s_{t+1}, a_{t+1})-\alpha \log \pi_\phi(a_{t+1}|s_{t+1}).
\end{equation}

SAC also utilizes target networks $Q_{\hat{\theta}_1}, Q_{\hat{\theta}_2}$ which are obtained as an Exponentially Moving Average (EMA) of the Q networks ($Q_{\theta_1}, Q_{\theta_2}$) for better learning stability, and gradient-based temperature tuning to determine the relative importance of the entropy,

\begin{equation}
    J(\alpha) = \mathbb{E}_{a_t\sim\pi_\phi(s_t)}[-\alpha \log \pi_\phi(a_t|s_t)-\alpha \bar{\mathcal{H}}].
\end{equation}

\subsection{Model-based RL : Dreamer}\label{sec:A.2}
To evaluate whether SNeRL can also be adopted in model-based RL algorithms, we adopt Dreamer \cite{hafner2019dream}. Dreamer learns the world model which consists of the following components:
\begin{equation}
\begin{split}
    &p_\theta(s_t|s_{t-1}, a_{t-1}, o_t)\\
    &q_\theta(o_t|s_t)\\
    &q_\theta(r_t|s_t)\\
    &q_\theta(s_t|s_{t-1}, a_{t-1}).
\end{split}
\end{equation}

These components are jointly optimized to increase the variational lower bound, which includes the following terms:
\begin{equation}
\begin{split}
    \mathcal{J}_O^t &=\ln q(o_t|s_t)\\
    \mathcal{J}_R^t &=\ln q(r_t|s_t)\\
    \mathcal{J}_D^t &= -\beta KL\big(p(s_t|s_{t-1}, a_{t-1}, o_{t}) || q(s_t|s_{t-1}, a_{t-1})\big).
\end{split}
\end{equation}

We replace the convolutional encoder of Dreamer with our feature extractor $\Omega$ (shared encoder of SNeRL) to design an RL with a dynamics model over the latent space with pre-trained mapping. To learn the action and value models, Dreamer optimizes the value model $v_\psi$ and the action model $q_\phi$ using the objectives
\begin{equation}
\begin{split}
    & \max_\phi \mathbb{E}\bigg(\sum_{\tau=t}^{t+H}V_\lambda(s_\tau)\bigg) \\
    & \max_\psi \mathbb{E}_{q_\theta, q_\phi}\bigg( \sum_{\tau=t}^{t+H} \frac{1}{2}\Big\vert\Big\vert  v_\psi(s_\tau)-V_\lambda(s_\tau) \Big\vert\Big\vert^2\bigg),
\end{split}
\end{equation}
where H denotes the horizon, and $V_\lambda$ the exponentially-weight value estimation. Refer to Dreamer \cite{hafner2019dream} for details.

\newpage
\subsection{Pseudo-code}\label{sec:A.3}
\begin{algorithm}
   \caption{Stage 1. pre-train multi-view encoder with SNeRL}
   \label{alg:overview}
\begin{algorithmic}[1]
    \STATE {\bfseries Input: encoder $\Omega$, off-the-shelf feature descriptor $\hat{F}$, offline dataset $\mathcal{D}$}
    \FOR{iteration=1,2,...,N}
        \FOR{sample minibatch $d$ from $\mathcal{D}$}
            \STATE $z \leftarrow \Omega (o^{1:V}, K^{1:V})$
            \STATE $\mathcal{L} \leftarrow \mathcal{L}_{\mathrm{RGB}} + \lambda_{sem}\mathcal{L}_{sem} + \lambda_{feat}\mathcal{L}_{feat}$
            \STATE update the parameters of $\Omega$ to minimize $\mathcal{L}$
        \ENDFOR
    \ENDFOR
\end{algorithmic}
\end{algorithm}

\begin{algorithm}[h!]
   \caption{Stage 2. Downstream Model-free RL (SAC)}
   \label{alg:overview}
\begin{algorithmic}[1]
    \STATE {\bfseries Input:} total training episodes $N$, Env, environment horizon $H$, actor network $\pi_\phi$, critic networks $Q_{\theta_{i=1,2}}$, target critic networks $Q_{\hat{\theta}_{i=1,2}}$, temperature $\alpha$, replay buffer $\mathcal{B}$, pre-trained encoder $\Omega$.
    \FOR{iteration=1,2,...,N}
        \STATE $o_{t=0} \leftarrow \mathtt{Env.reset()}$
        \FOR{$t$=0,1,...,$H$-1}
            \STATE $a_t \leftarrow \pi_\phi(\cdot \vert \Omega(o_t))$
            \STATE $r_{t}, o_{t+1}\leftarrow\mathtt{Env.step}(a_t)$ 
            \STATE $\mathcal{B} \leftarrow \mathcal{B}\cup\{(o_t, a_t, r_t, o_{t+1})\}$
        \ENDFOR
        \FOR {each gradient step}
            \STATE $\theta_i \leftarrow{\theta_i} - \lambda_{Q}\nabla_{\theta_i}J_Q(\theta_i)$
            \STATE $\phi \leftarrow{\phi} - \lambda_{\pi}\nabla_{\phi}J_\pi(\phi)$
            \STATE $\alpha \leftarrow{\alpha} - \lambda\nabla_{\alpha}J(\alpha)$
            \IF{update target critic networks}
                \STATE $\hat{\theta}_i \leftarrow{\tau{\theta}_{i}}+(1-\tau)\hat{\theta}_{i}$
            \ENDIF
        \ENDFOR
    \ENDFOR
\end{algorithmic}
\end{algorithm}

\begin{algorithm}[h!]
   \caption{Stage 2. Downstream Model-based RL (Dreamer)}
   \label{alg:overview}
\begin{algorithmic}[1]
    \STATE {\bfseries Input:} total training episodes $N$, update step $C$, Env, environment horizon $T$, imagination horizon $H$, Neural network parameters $\theta, \phi, \psi$, replay buffer $\mathcal{B}$, pre-trained encoder $\Omega$.
    \FOR{iteration=1,2,...,N}
        \FOR{$c$=1,...,$C$}
            \STATE sample data sequence $\{(a_t, o_t, r_t)\}_{t=0,...,H-1}\sim\mathcal{B}$
            \STATE compute model states $s_t \sim p_\theta(s_t|\Omega(o_{t-1}), a_{t-1})$ and reward $q_\theta(r_t|\Omega(o_t))$
            \STATE update $\theta$ using representation learning
            \STATE imaging trajectories and compute value estimates $V_\lambda(s_\tau)$
            \STATE $\phi \leftarrow \phi + \alpha \nabla_\phi\sum_{\tau=t}^{t+H} V_\lambda (s_\tau)$
            \STATE $\psi \leftarrow \psi + \alpha \nabla_\psi\sum_{\tau=t}^{t+H} \frac{1}{2}|| v_\psi(s_\tau)-V_\lambda (s_\tau)||^2$
        \ENDFOR
        \STATE $o_{t=0} \leftarrow \mathtt{Env.reset()}$
        \FOR{$t$=0,1,...,$T-1$}
            \STATE $a_t \leftarrow q_\phi(\cdot \vert \Omega(o_t))$
            \STATE $r_{t}, o_{t+1}\leftarrow\mathtt{Env.step}(a_t)$ 
            \STATE $\mathcal{B} \leftarrow \mathcal{B}\cup\{(o_t, a_t, r_t, o_{t+1})\}$
        \ENDFOR
    \ENDFOR
\end{algorithmic}
\end{algorithm}
\newpage


\section{Training \& Experiments Details}

\subsection{Encoder Architecture}
We design the encoder architecture similar to \citet{laskin2020curl}, which consists of multiple convolutional layers and ReLU activation, but modify it to be applicable in the multi-view observation inputs.
The same encoder is also adopted in the actor and critic to embed the pixel-level (multi-view) observations.
We demonstrate the details of the convolutional encoder with PyTorch-like pseudo-code as below.

\begin{algorithm}
\caption{Multi-view Encoder Pseudocode, PyTorch-like}
\label{alg:code}
\definecolor{codeblue}{rgb}{0.25,0.5,0.5}
\definecolor{codekw}{rgb}{0.85, 0.18, 0.50}
\lstset{
  backgroundcolor=\color{white},
  basicstyle=\fontsize{7.5pt}{7.5pt}\ttfamily\selectfont,
  columns=fullflexible,
  breaklines=true,
  captionpos=b,
  commentstyle=\fontsize{7.5pt}{7.5pt}\color{codeblue},
  keywordstyle=\fontsize{7.5pt}{7.5pt}\color{codekw},
}
\begin{lstlisting}[language=python]
def encoder(x1, x2, x3, K1, K2, K3, z_dim):
    """
    Multi-view ConvNet encoder
    args:
        B = batch_size, C = channels,
        H, W =spatial_dims
        x1, x2, x3: images from 3 different camera views
        x1, x2, x3 shape: [B, C, H, W]
        K1, K2, K3: camera poses from 3 different camera views
        K1, K2, K3 shape: [B, 4, 4]
        z_dim = latent dimension
    """
    x = x / 255.

    # c: channels, f: filters
    # k: kernel, s: stride

    z1 = Conv2d(c=x1.shape[1], f=32, k=3, s=2)(x1)
    z1 = ReLU(z1)

    z2 = Conv2d(c=x2.shape[1], f=32, k=3, s=2)(x2)
    z2 = ReLU(z2)

    z3 = Conv2d(c=x3.shape[1], f=32, k=3, s=2)(x3)
    z3 = ReLU(z3)

    for _ in range(num_layers-1):
    
        z1 = Conv2d(c=32, f=32, k=3, s=1)(z1)
        z1 = ReLU(z1)

        z2 = Conv2d(c=32, f=32, k=3, s=1)(z2)
        z2 = ReLU(z2)

        z3 = Conv2d(c=32, f=32, k=3, s=1)(z3)
        z3 = ReLU(z3)

    z1 = flatten(z1)
    z2 = flatten(z2)
    z3 = flatten(z3)

    z1 = concat([z1, K1.view(B,16)], dim=1)
    z2 = concat([z2, K1.view(B,16)], dim=1)
    z3 = concat([z3, K1.view(B,16)], dim=1)

    z1 = Linear(z1.shape[1], z_dim)(z1)
    z2 = Linear(z2.shape[1], z_dim)(z2)
    z3 = Linear(z3.shape[1], z_dim)(z3)

    z = concat([z1, z2, z3], dim=1).mean(dim=1)

    z = Linear(z.shape[1], z_dim)(z)
    z = LayerNorm(z)
    z = tanh(z)

    return z
\end{lstlisting}
\end{algorithm}
\newpage
\subsection{Datasets}\label{sec:B.2}
The offline datasets for SNeRL and baselines consist of 14400 scenes. Each scene consists of three image observations taken from different camera views. The observations from each camera view are represented in Figure \ref{fig:thumbnail} (Window-open-v2) and Figure \ref{fig:recon} (Soccer-v2, Hammer-v2, Drawer-open-v2). To collect the dataset, we utilized random actions and the policies provided by Meta-world (half-and-half mixed).

\begin{figure*}[h!]
    \centering
    \includegraphics[width=0.995\textwidth]{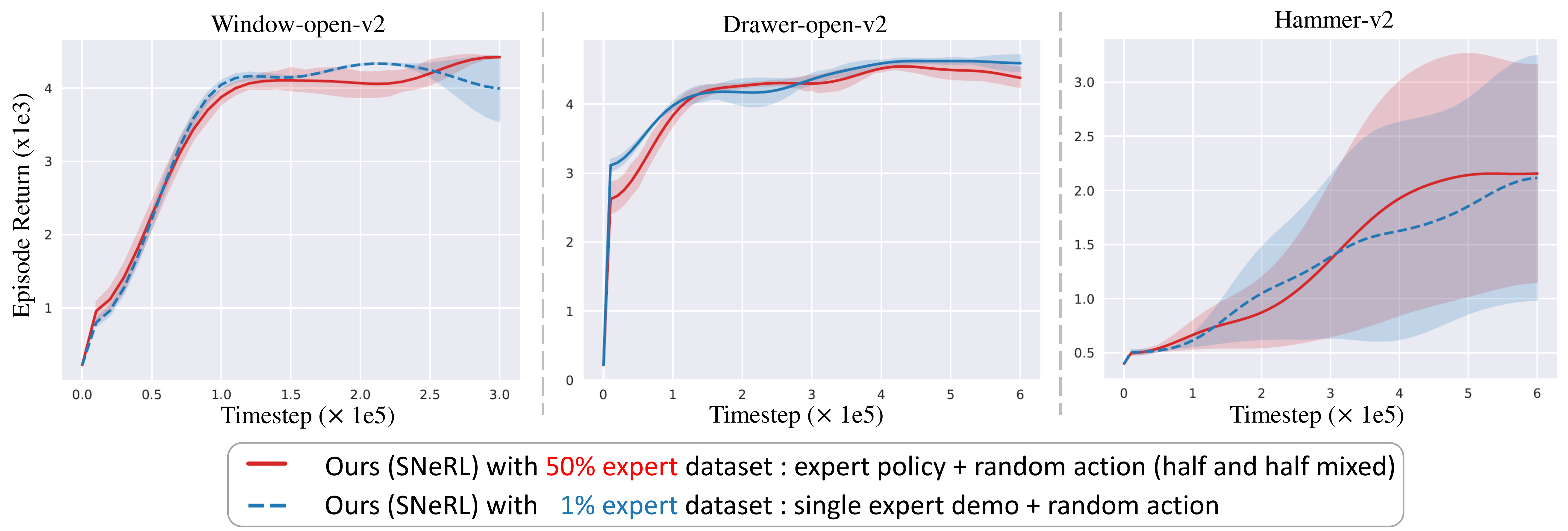}
    \vspace{-12pt}
    \caption{The performance of SNeRL with different quality of offline datasets.
    }
    \vspace{-8pt}
    \label{fig:single_demo}
    
\end{figure*}

To observe how the performance of the proposed method varies with the quality of the dataset, we further trained the SNeRL encoder with a dataset collected by a single expert demo and random actions. Only 120 scenes of the total scenes (14400, 120/14400$\simeq$1\%) were obtained from the path of the expert demo, and the remaining 14280 scenes were obtained by taking random actions from one moment of the path of the expert demo. As shown in Figure \ref{fig:single_demo}, we observe that the quality of the dataset slightly affects the learning stability, but there is no dramatic performance degradation. The results show that there would be no significant degradation in the performance of the SNeRL if the dataset adequately covers the state space, even if the policy that collects the offline dataset is suboptimal.

\subsection{Computational Resources}
Stage 1 (pre-training encoder) in our experiments has been performed using a single NVIDIA RTX A6000 and AMD Ryzen 2950X, and stage 2 (RL downstream tasks) has been performed using an NVIDIA RTX A5000 and AMD Ryzen 2950X. 
Training the SNeRL encoder takes 4-5 days and learning model-based RL and model-free RL takes 1-2 days.

\subsection{Hyperparameters}
\begin{table}[h]
\centering
\caption{Hyperparameters for pre-training multi-view encoder}
\label{table:hyperparameter_stage1}
\begin{tabular}{c|c}
& SNeRL \\
\hline
Convolution layers & 4\\
Number of filters & 32\\
Non-linearity & ReLU\\
MLP layers for NeRF & 8\\
Hidden units (MLP) & 256\\
Number of different views & 3\\
NeRF learning rate & 5e-4\\
Number of rays per gradient step & 1024\\
Number of samples per ray&64\\
\end{tabular}
\end{table}

\begin{table}[h]
\centering
\caption{Hyperparameters for SAC (for SNeRL and baselines)}
\label{table:hyperparameter_stage2}
\begin{tabular}{c|c}
& SAC \\
\hline
hidden layer & (1024, 1024)\\
frame stack & 2 \\
replay buffer size & 100000 \\
initial random steps & 1000 \\
batch size & 128 \\
actor learning rate & 1e-3 \\
critic learning rate & 1e-3\\
$\alpha$ learning rate & 1e-4\\
$\beta$ for Adam optimizer (actor, critic) & 0.9\\
eps for Adam optimizer ($\alpha$) & 1e-08\\
$\beta$ for Adam optimizer ($\alpha$) & 0.5\\
eps for Adam optimizer ($\alpha$) & 1e-08\\
critic target update interval & 2 \\
actor network update interval & 2 \\
actor log std min, max & -10, 2\\
init temperature & 0.1\\
$\tau$ for EMA & 0.01\\
discount factor $\gamma$ & 0.99\\
\end{tabular}
\end{table}

\begin{table}[h]
\centering
\caption{Hyperparameters for Dreamer (for SNeRL and baselines)}
\label{table:hyperparameter_stage2}
\begin{tabular}{c|c}
& Dreamer \\
\hline
embedding size & 63 \\
hidden / belief size & 128\\
state size & 30 \\
action noise & 0.3 \\
batch size & 32 \\
world model learning rate & 1e-3 \\
actor learning rate & 5e-5 \\
value network learning rate & 5e-5 \\
discount factor $\gamma$ & 0.99\\
replay buffer size & 100000 \\
planning horizon & 15 \\
eps for Adam optimizer & 1e-07\\

\end{tabular}
\end{table}


\end{document}